# KINEMATIC ANALYSIS OF GEARED ROBOTIC MECHANISM USING MATROID AND T-T GRAPH METHODS.

Seyedvahid Amirinezhad Mustafa K. Uyguroğlu

Department of Electrical and Enectronic Engineering
Faculty of Engineering
Eastern Mediterranean University
Gazimağusa –North Cyprus
Vahid.amirinezhad@cc.emu.edu.tr
Mustafa.uyguroglu@emu.edu.tr

## *A. ABSTRACT*

In this paper, the kinematic structure of the geared robotic mechanism (GRM) is investigated with the aid of two different methods which are based on directed graphs and the methods are compared. One of the methods is Matroid Method developed by Talpasanu and the other method is Tsai-Tokad (T-T) Graph method developed by Uyguroglu and Demirel. It is shown that the kinematic structure of the geared robotic mechanism can be represented by directed graphs and angular velocity equations of the mechanisms can be systematically obtained from the graphs. The advantages and disadvantages of both methods are demonstrated relative to each other.

## *B. INTRODUCTION*

Kinematic and dynamic analysis of mechanical systems have been well established by using graph theory in recent years. Non-oriented and oriented graphs were used for this purpose. Non-oriented graph technique is mainly used for the kinematic analysis of robotic bevel-gear trains [1]–[4]. The oriented graph technique has been used for electrical circuits and other types of lumped physical systems including mechanical systems in one-dimensional motion since the early sixties [5]–[8]. Chou et al. [9] used these techniques to three-dimensional systems. Recently, Tokad developed a systematic approach, the so called Network Model Approach, for the formulation of three dimensional mechanical systems [10], and Uyguroglu and Tokad extended this approach for the kinematic and dynamic analysis of spatial robotic bevel-gear trains [11]. Most recently the oriented and non-oriented graph techniques were compared by Uyguroglu and Demirel [12], and the advantages of the oriented graph over the non-oriented graph were demonstrated using the kinematic analysis of bevel-gear trains. In order to overcome the weaknesses of the methods developed by Tsai and Tokad, both method were combined and the so called T-T Graph method was introduced []. On the other hand, Talpasanu et al. [] developed Matroid Method for the kinematic analysis of geared mechanisms based on directed graph as well.

In this paper the kinematic structures of the GRM is investigated with the aid of Matroid Method and T-T Graph method. Since both methods are used directed graph, the similarities and the differences are shown and the advantages of each system are indicated.

### *C.* ***Geared Robotic Mechanism***

GRMs are closed-loop configurations which are used to reduce the mass and inertia of the actuators' loads. Gear trains in GRMs are employed such that actuators can be placed as closely as possible to the base. Figure 1 shows functional schematic of the GRM. It has 3 Degree of Freedom (i.e. it has 3 inputs) and the end-effector can have spatial motion (3 Dimensional) because this mechanism can generate two rotations about two intersected axes and one rotation of end-effector about its axis. In this mechanism 4, 5, and 6 are sun gears (input links), 1 and 2 are carriers and $3, 7', \text{and } 7''$ are planet gears. It is observed that links and joints (gear train) are used to transmit the rotation of the inputs to the end-effector. The motion of end-effector is produced by links 4, 5, and 6 as inputs. The end-effector is attached to link 3 and carried by link 2. *M1, M2 and M3* are actuators. The rotation of link 3 is caused by *M3* through 6 and 7 links and the rotation of links 1 and 2 is made by *M1* through link 4 and *M2* through link 5 respectively.

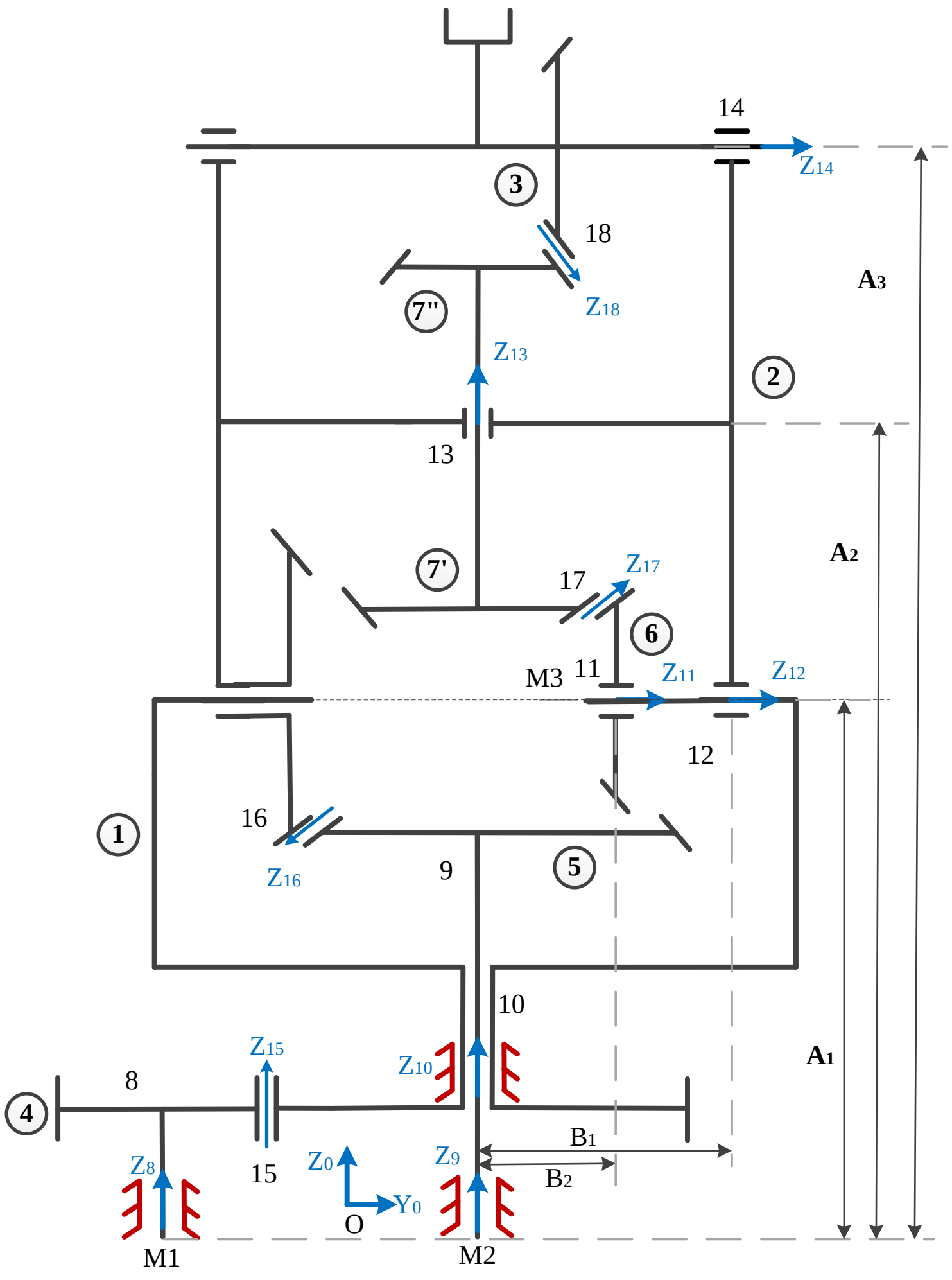

Figure 1. The GRM mechanism.

## II. MATROID METHOD

In this section, Matroid method […] is applied to the sample Geared Robotics Mechanism (GRM) to obtain the kinematic equations. First, its digraph is sketched and corresponding matrices are obtained. Then, we calculate relative angular velocities by using these matrices and Screw theory.

**Associated digraph and corresponding matrices:**

The mechanism in Figure 1 consists of $n = 7$ links, $t = 7$ turning pairs and $c = 4$ gear pairs. Note that $k = 11$ is total number of joints (i.e. $k = t + c$ and $t = n$ ). The following labeling, which is assigned to links and joints of sample mechanism, is used in Matroid method […]:

- 0 is assigned to ground link.
- 1, 2, 3, 4, 5, 6, and 7 are assigned to gears and carriers.
- 8, 9, 10, 11, 12, 13, and 14 are assigned to turning joints.
- 15, 16, 17, and 18 are assigned to meshing joints.

Figure 2 shows associated digraph of the sample mechanism. In this digraph, nodes present links as well as solid and dash arrows indicate turning and meshing joints respectively. Note that in each mechanism, corresponding to *c* gear pair there exist *c fundamental cycles* hence in sample mechanism, there are 4 fundamental cycles: $C_{15}, C_{16}, C_{17}, \text{ and } C_{18}$. In addition, *Spanning Tree* is defined such that there is not any cycle in digraph so in Figure 2 collection of solid arrows creates Spanning Tree.

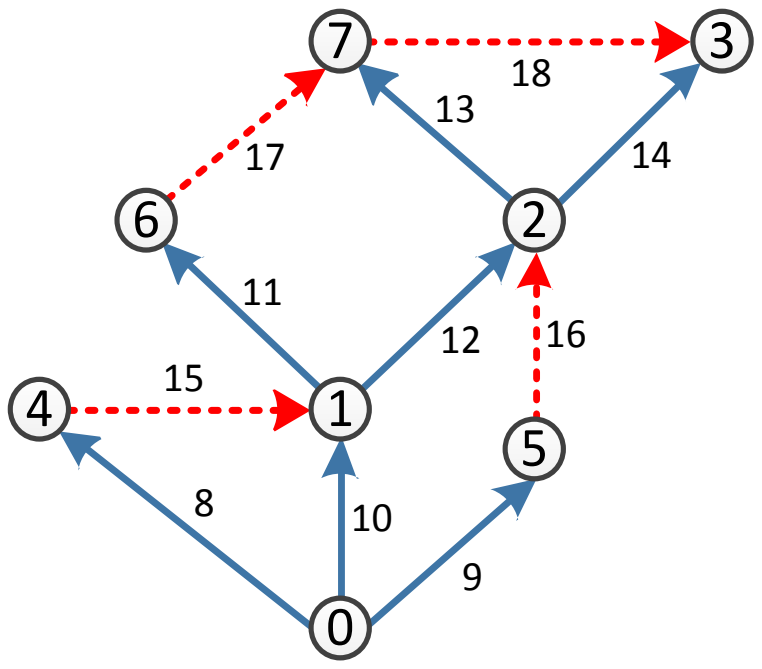

Figure 2. Mechanism associated digraph.

Here, one could easily obtain the Incidence Node-Edge matrix from Figure 2:

$$
\mathbf{\Gamma}^{0} = \begin{array}{c} \\ \\ 0\rightarrow \\ 1\rightarrow \\ 2\rightarrow \\ 3\rightarrow \\ 4\rightarrow \\ 5\rightarrow \\ 6\rightarrow \\ 7\rightarrow \end{array}
\begin{array}{cccccc:ccccc}
8 & 9 & 10 & 11 & 12 & 13 & 14 & 15 & 16 & 17 & 18 \\
\downarrow & \downarrow & \downarrow & \downarrow & \downarrow & \downarrow & \downarrow & \downarrow & \downarrow & \downarrow & \downarrow \\
-1 & -1 & -1 & 0 & 0 & 0 & 0 & 0 & 0 & 0 & 0 \\
\hdashline
0 & 0 & 1 & -1 & -1 & 0 & 0 & 1 & 0 & 0 & 0 \\
0 & 0 & 0 & 0 & 1 & -1 & -1 & 0 & 1 & 0 & 0 \\
0 & 0 & 0 & 0 & 0 & 0 & 1 & 0 & 0 & 0 & 1 \\
1 & 0 & 0 & 0 & 0 & 0 & 0 & -1 & 0 & 0 & 0 \\
0 & 1 & 0 & 0 & 0 & 0 & 0 & 0 & -1 & 0 & 0 \\
0 & 0 & 0 & 1 & 0 & 0 & 0 & 0 & 0 & -1 & 0 \\
0 & 0 & 0 & 0 & 0 & 1 & 0 & 0 & 0 & 1 & -1
\end{array} \tag{1}
$$

The entries of an Incident Node-Edge Matrix are 1, -1 and 0. Each column represents an directed edge which connects two nodes and contains two nonzero entries. The arrow head side is 1 and the other side is -1. Reduced Incident Node-Edge matrix $\mathbf{\Gamma}$ is obtained by deleting the first row where rows of this matrix are independent. Since $k = t + c$, $\mathbf{\Gamma}$ consists of two sub-matrices: $\mathbf{G}$ and $\mathbf{G}^{*}$ which correspond to turning and gear pairs respectively.

$$\boldsymbol{\Gamma} = \left[ \mathbf{G} \;\vdots\; \mathbf{G}^{*} \right] = \left[ \begin{array}{ccccccc:cccc} 0 & 0 & 1 & -1 & -1 & 0 & 0 & 1 & 0 & 0 & 0 \\ 0 & 0 & 0 & 0 & 1 & -1 & -1 & 0 & 1 & 0 & 0 \\ 0 & 0 & 0 & 0 & 0 & 0 & 1 & 0 & 0 & 0 & 1 \\ 1 & 0 & 0 & 0 & 0 & 0 & 0 & -1 & 0 & 0 & 0 \\ 0 & 1 & 0 & 0 & 0 & 0 & 0 & 0 & -1 & 0 & 0 \\ 0 & 0 & 0 & 1 & 0 & 0 & 0 & 0 & 0 & -1 & 0 \\ 0 & 0 & 0 & 0 & 0 & 1 & 0 & 0 & 0 & 1 & -1 \end{array} \right] \quad (2).$$

Now, regarding to associated digraph in Figure 2, we can obtain Path matrix as follows:

$$\mathbf{Z} = \begin{array}{c} \\ \\ 8 \rightarrow \\ 9 \rightarrow \\ 10 \rightarrow \\ 11 \rightarrow \\ 12 \rightarrow \\ 13 \rightarrow \\ 14 \rightarrow \end{array} \begin{array}{ccccccc} 1 & 2 & 3 & 4 & 5 & 6 & 7 \\ \downarrow & \downarrow & \downarrow & \downarrow & \downarrow & \downarrow & \downarrow \\ \left[ 0 \right. & 0 & 0 & -1 & 0 & 0 & \left. 0 \right] \\ 0 & 0 & 0 & 0 & -1 & 0 & 0 \\ -1 & -1 & -1 & 0 & 0 & -1 & -1 \\ 0 & 0 & 0 & 0 & 0 & -1 & 0 \\ 0 & -1 & -1 & 0 & 0 & 0 & -1 \\ 0 & 0 & 0 & 0 & 0 & 0 & -1 \\ 0 & 0 & -1 & 0 & 0 & 0 & 0 \end{array} \quad (3).$$

Path matrix $\mathbf{z}$ [...], is a $t \times n$ matrix and comes from the spanning tree. Here, $z_{t,n}$ (the entries of Path matrix) can be -1, 0, and +1. If edge *t* belongs to one of Spanning Tree's paths which are started from node *n* toward the ground link and its orientation is the same as path's direction, $z_{t,n}$ becomes +1. If it belongs to the path but the orientation is opposite, $z_{t,n}$ becomes -1. $z_{t,n} = 0$ if edge does not belong to the related path.

Spanning Tree Matrix is obtained by using the formula:

$$\mathbf{T} = \mathbf{G}^{*T} \cdot \mathbf{Z}^{T} \quad (4).$$

Then, Cycle-Basis matrix is obtained as:

$$
\mathbf{C} = \left[\mathbf{T} \;\vdots\; \mathbf{U}\right] =
\begin{array}{c}
\\
\\
C_{15} \rightarrow \\
C_{16} \rightarrow \\
C_{17} \rightarrow \\
C_{18} \rightarrow
\end{array}
\begin{array}{c}
\begin{array}{ccccccccccc}
8 & 9 & 10 & 11 & 12 & 13 & 14 & 15 & 16 & 17 & 18 \\
\downarrow & \downarrow & \downarrow & \downarrow & \downarrow & \downarrow & \downarrow & \downarrow & \downarrow & \downarrow & \downarrow
\end{array} \\
\left[\begin{array}{ccccccc:cccc}
1 & 0 & -1 & 0 & 0 & 0 & 0 & 1 & 0 & 0 & 0 \\
0 & 1 & -1 & 0 & -1 & 0 & 0 & 0 & 1 & 0 & 0 \\
0 & 0 & 0 & 1 & -1 & -1 & 0 & 0 & 0 & 1 & 0 \\
0 & 0 & 0 & 0 & 0 & 1 & -1 & 0 & 0 & 0 & 1
\end{array}\right]
\end{array}
\tag{5}
$$

Actually, one can obtain Cycle-Basis matrix from Figure 2 directly.

1) Independent equations for relative angular velocities:

In this part, we obtain relative angular velocity equations of joints by means of Screw theory and then define output relative velocities in terms of inputs. Let consider the dual vector (Screw) in Eq. (6):

$$
\hat{\mathbf{u}}^{0}_{c,k} = \left(\frac{\mathbf{u}^{0}_{k}}{\mathbf{r}^{0}_{c,k}}\right) = \left(L_k \quad M_k \quad N_k \;\middle|\; P_{c,k} \quad Q_{c,k} \quad R_{c,k}\right)^{T} \tag{6}
$$

This $6 \times 1$ matrix can define spatial geometry of z-axes of local frames attached to *k* joints. Note that each of these local frames has unit vector $\mathbf{u} = \left(0 \quad 0 \quad 1\right)^{T}$ with respect to itself. The first vector of Screw is unit vector of $z_k$ axis with respect to the base frame and it presents the orientation of $z_k$ axis:

$$
\mathbf{u}^{0}_{k} = \mathbf{D}_{0,k} \cdot \mathbf{u} \tag{7}
$$

where $\mathbf{D}_{0,k}$ is a pure rotational matrix about $x_k$ axis:

$$
\mathbf{D}_{0,k} = \begin{bmatrix} 1 & 0 & 0 \\ 0 & \cos\varphi_k & -\sin\varphi_k \\ 0 & \sin\varphi_k & \cos\varphi_k \end{bmatrix} \tag{8}
$$

So components of $\mathbf{u}_k^0$ are:

$$L_k = 0 \; ; \; M_k = -\sin\varphi_k \; ; \; N_k = \cos\varphi_k \tag{9}$$

where $\varphi_k$ are offset angles between z-axis of base and z-axes of turning axes. So, in sample mechanism: $\varphi_8 = \varphi_9 = \varphi_{10} = \varphi_{13} = 0$ and $\varphi_{11} = \varphi_{12} = \varphi_{14} = -90°$ and unit vectors of revolute joints are: $\mathbf{u}_8^0 = \mathbf{u}_9^0 = \mathbf{u}_{10}^0 = \mathbf{u}_{13}^0 = \begin{pmatrix} 0 & 0 & 1 \end{pmatrix}^T$ and $\mathbf{u}_{11}^0 = \mathbf{u}_{12}^0 = \mathbf{u}_{14}^0 = \begin{pmatrix} 0 & 1 & 0 \end{pmatrix}^T$. The second vector is the position vector of with respect to the reference frame:

$$\mathbf{r}_{c,k}^0 = \mathbf{I}_{c,k}^0 \times \mathbf{u}_k^0 \tag{10}$$

where $\mathbf{I}_{c,k}^0 = \begin{pmatrix} x_{c,k} & y_{c,k} & z_{c,k} \end{pmatrix}^T$ is distance vector which orients from *c* to *k* and it has the form:

$$\begin{pmatrix} x_k - x_c & y_k - y_c & z_k - z_c \end{pmatrix}^T \qquad (11).$$

According to Table 1, components of position vector are:

$$P_{c,k} = z_{c,k}\sin\varphi_k + y_{c,k}\cos\varphi_k \; ; Q_{c,k} = 0 \; ; \; R_{c,k} = 0 \qquad (12).$$

Table 1. Coordinates of turning and gear pairs

| | 8 | 9 | 10 | 11 | 12 | 13 | 14 | 15 | 16 | 17 | 18 |
|---|---|---|---|---|---|---|---|---|---|---|---|
| $x_k$ | 0 | 0 | 0 | 0 | 0 | 0 | 0 | 0 | 0 | 0 | 0 |
| $y_k$ | $-(d_1+d_4)/2$ | 0 | 0 | $B_2$ | $B_1$ | 0 | $B_1$ | $-d_1/2$ | $-d_5/2$ | $d'_7/2$ | $d''_7/2$ |
| $z_k$ | 0 | 0 | $A_1 - d_2/2$ | $A_1$ | $A_1$ | $A_2$ | $A_3$ | 0 | $A_1 - d_2/2$ | $A_1 + d_6/2$ | $A_3 - d_3/2$ |

So for each cycle, the $P_{c,k}$ coefficients are calculated as follows:

Cycle $C_{15}$ : $P_{15,8} = -\dfrac{d_4}{2}$ ; $P_{15,10} = \dfrac{d_1}{2}$ ; $P_{15,15} = 0$ .

Cycle $C_{16}$ : $P_{16,9} = \frac{d_5}{2}$ ; $P_{16,10} = \frac{d_5}{2}$ ; $P_{16,12} = -\frac{d_2}{2}$ ; $P_{16,16} = 0$ .

Cycle $C_{17}$ : $P_{17,11} = \frac{d_6}{2}$ ; $P_{17,12} = \frac{d_6}{2}$ ; $P_{17,13} = -\frac{d_7'}{2}$ ; $P_{17,17} = 0$ .

Cycle $C_{18}$ : $P_{18,13} = -\frac{d_7''}{2}$ ; $P_{18,14} = -\frac{d_3}{2}$ ; $P_{18,18} = 0$ .

Here, we define *twist* matrix as a product between screw and relative velocity variables of pairs [...]:

$$\hat{\mathbf{s}}_k^0 = \hat{\mathbf{u}}_{c,k}^0 \cdot \dot{\boldsymbol{\theta}}_k \qquad (13)$$

where $\dot{\boldsymbol{\theta}}_k$ is partitioned into two sub-matrices, $\dot{\boldsymbol{\theta}}_t = \begin{pmatrix} \dot{\theta}_8 & \dot{\theta}_9 & \dot{\theta}_{10} & \dot{\theta}_{11} & \dot{\theta}_{12} & \dot{\theta}_{13} & \dot{\theta}_{14} \end{pmatrix}^T$ and $\dot{\boldsymbol{\theta}}_c = \begin{pmatrix} \dot{\theta}_{15} & \dot{\theta}_{16} & \dot{\theta}_{17} & \dot{\theta}_{18} \end{pmatrix}^T$ relates to velocities of turning and gear pairs respectively. Now, we can obtain relative angular velocity equations by applying Hadamard entry-wise product on Eq. (5) and Eq. (13) as follows:

$$[\mathbf{C} \circ \hat{\mathbf{s}}_k^0] = \mathbf{0}_c \qquad (14)$$

where $\hat{\mathbf{s}}_k^0 = \left[ \hat{\mathbf{s}}_t^0 \;\vdots\; \hat{\mathbf{s}}_c^0 \right]$. In [...], to two orthogonality conditions, one for relative velocities and another for relative moments, are defined in order to Eq. (14) holds true. According to these two conditions, we can express equations of relative velocity variables of meshing and turning pairs in Eq. (15) and Eq. (16) respectively:

$$\dot{\boldsymbol{\theta}}_c = -\left[ \mathbf{T} \circ \mathbf{u}_t^0 \right] \cdot \dot{\boldsymbol{\theta}}_t \qquad (15)$$

and because $\mathbf{r}^0_{c,c} = \mathbf{0}$ then

$$\left[\mathbf{T} \circ \mathbf{r}^0_{c,t}\right] \cdot \dot{\boldsymbol{\theta}}_t = \mathbf{0}_c \tag{16}$$

Since $Q_{c,t} = R_{c,t} = 0$, we can simplify Eq. (16) in the following form:

$$\left[\mathbf{P}_{c,t}\right] \cdot \dot{\boldsymbol{\theta}}_t = \mathbf{0}_c \tag{17}$$

where $\mathbf{P}_{c,t}$ is coefficient matrix:

$$\mathbf{P}_{c,t} = \left[\mathbf{T} \circ P_{c,t}\right] \tag{18}$$

In Eq. (17), $P_{c,t}$ are coefficients in terms of pitch diameter. These scalar coefficients are used to acquire independent equations of relative angular velocities. For sample mechanism, these independent equations are expressed in Eq. (19):

$$\begin{bmatrix} P_{15,8} & 0 & -P_{15,10} & 0 & 0 & 0 & 0 \\ 0 & P_{16,9} & -P_{16,10} & 0 & -P_{16,12} & 0 & 0 \\ 0 & 0 & 0 & P_{17,11} & -P_{17,12} & -P_{17,13} & 0 \\ 0 & 0 & 0 & 0 & 0 & P_{18,13} & -P_{18,14} \end{bmatrix} \cdot \begin{pmatrix} \dot{\theta}_8 \\ \dot{\theta}_9 \\ \dot{\theta}_{10} \\ \dot{\theta}_{11} \\ \dot{\theta}_{12} \\ \dot{\theta}_{13} \\ \dot{\theta}_{14} \end{pmatrix} = \begin{pmatrix} 0 \\ 0 \\ 0 \\ 0 \end{pmatrix} \tag{19}$$

Moreover, we will later rewrite these coefficients in terms of gear ratios in Eq. (24):

$$i_{15} = \frac{d_4}{d_1};\, i_{16} = \frac{d_5}{d_2};\, i_{17} = \frac{d_6}{d_7'};\, i_{18} = \frac{d_7''}{d_3} \tag{20}$$

Up to here, we obtain a set of independent equations for relative angular velocities of turning pairs including input pairs and output ones. Now according to Kutzbach criterion […] in Eq. (21) , we can calculate output relative velocities in terms of input relative velocities:

$$E = t - r \tag{21}$$

where $E$ is the number of inputs (Degree of Freedom), $t$ is the number of turning pairs and $r$ is the number of outputs (rank of Cycle-Basis matrix). So Eq. (17) is partitioned in the following form:

$$\left[ \mathbf{P}_{r,E} \; \vdots \; \mathbf{P}_{r,r} \right] \cdot \begin{pmatrix} \dot{\boldsymbol{\theta}}_E \\ \hdashline \dot{\boldsymbol{\theta}}_r \end{pmatrix} = \mathbf{0} \tag{22}$$

Hence, solutions for output relative velocities $\dot{\boldsymbol{\theta}}_r$ can be defined as functions of input relative velocities $\dot{\boldsymbol{\theta}}_E$ :

$$\left( \dot{\boldsymbol{\theta}}_r \right) = -\left[ \mathbf{P}_r \right]^{-1} \cdot \left[ \mathbf{P}_E \right] \cdot \left( \dot{\boldsymbol{\theta}}_E \right) \tag{23}$$

For sample mechanism in Figure 1, since $E = 3,\ t = 7 \text{ and } r = 4$ so $\dot{\theta}_8, \dot{\theta}_9 \text{ and } \dot{\theta}_{11}$ are input velocities and $\dot{\theta}_{10}, \dot{\theta}_{12}, \dot{\theta}_{13} \text{ and } \dot{\theta}_{14}$ are output velocities. According to Eq. (20) and Eq. (22), we can rewrite Eq. (19) as follows:

$$\begin{bmatrix} -i_{15} & 0 & 0 & -1 & 0 & 0 & 0 \\ 0 & +i_{16} & 0 & -i_{16} & +1 & 0 & 0 \\ 0 & 0 & +i_{17} & 0 & -i_{17} & +1 & 0 \\ 0 & 0 & 0 & 0 & 0 & -i_{18} & +1 \end{bmatrix} \cdot \begin{pmatrix} \dot{\theta}_8 \\ \dot{\theta}_9 \\ \dot{\theta}_{11} \\ \hdashline \dot{\theta}_{10} \\ \dot{\theta}_{12} \\ \dot{\theta}_{13} \\ \dot{\theta}_{14} \end{pmatrix} = \begin{pmatrix} 0 \\ 0 \\ 0 \\ 0 \end{pmatrix} \tag{24}$$

Note that in Eq. (24), we changed the order of third and fourth columns in $\mathbf{P}$ matrix and third and fourth rows in $\dot{\boldsymbol{\theta}}$ vector compared with Eq. (19).

According to Eq. (23), we can obtain independent equations for output relative velocities in terms of gear ratio for the sample mechanism:

$$\begin{pmatrix} \dot{\theta}_{10} \\ \dot{\theta}_{12} \\ \dot{\theta}_{13} \\ \dot{\theta}_{14} \end{pmatrix} = \begin{bmatrix} -i_{15} & 0 & 0 \\ -i_{16} i_{15} & -i_{16} & 0 \\ -i_{17} i_{16} i_{15} & -i_{17} i_{16} & -i_{17} \\ -i_{18} i_{17} i_{16} i_{15} & -i_{18} i_{17} i_{16} & -i_{18} i_{17} \end{bmatrix} \cdot \begin{pmatrix} \dot{\theta}_{8} \\ \dot{\theta}_{9} \\ \dot{\theta}_{11} \end{pmatrix} \tag{25}$$

## III. TSAI–TOKAD (T–T) GRAPH METHOD

On the functional schematic of the GRM shown in Figure 3:

- 0 is assigned to the ground.
- Links are numbered as 1, 2, 3, 4, 5, 6, and 7.
- Turning pairs’ axes are labeled as a,b,c,d, and e.

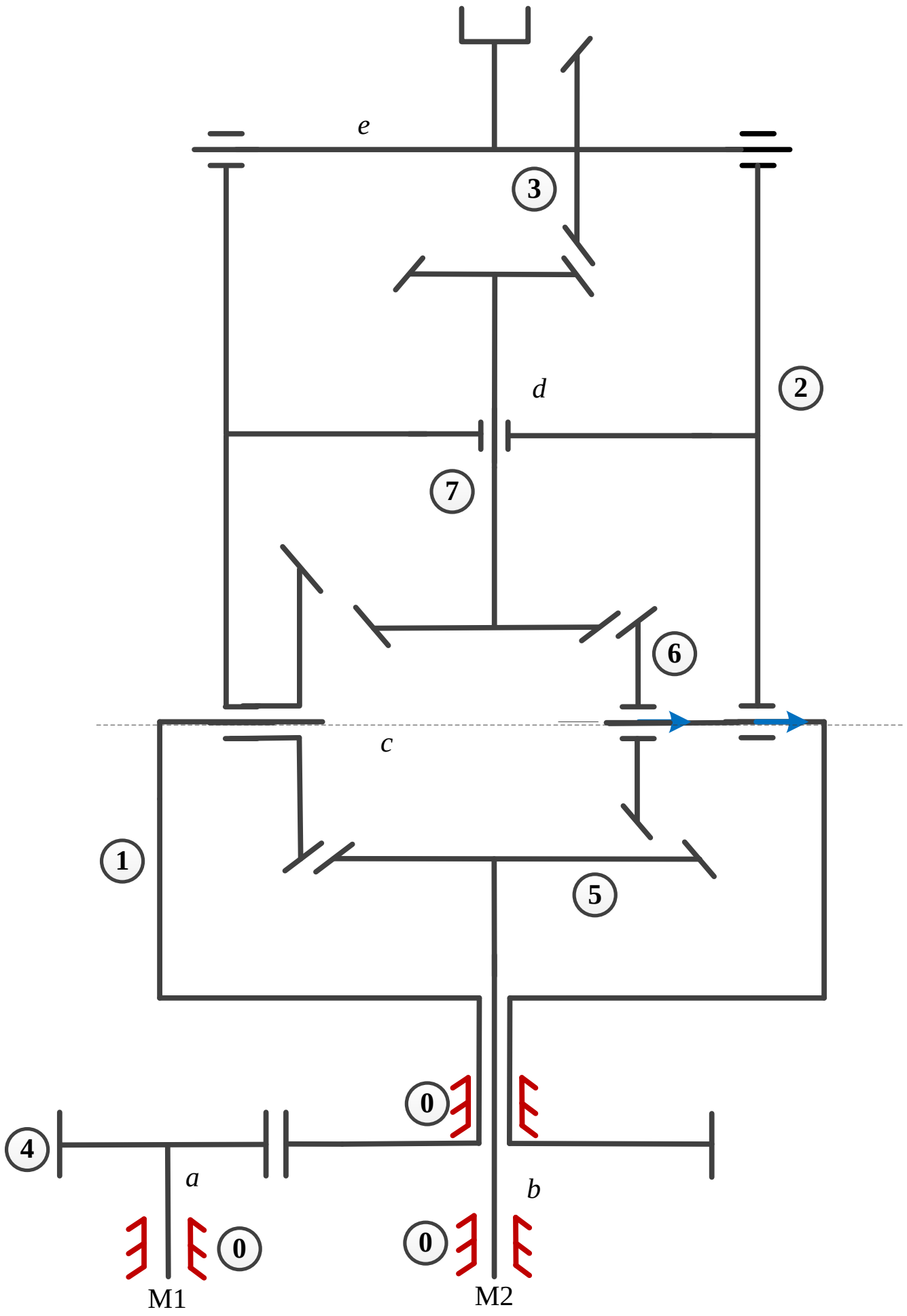

Figure 3: Functioal Schematic of the GRM.

**T-T Graph Representing  of the GRM**

In the T-T graph method, links are represented by nodes and the oriented lines between these nodes indicate the terminal pairs (ports), where a pair of meters, real or conceptual, are connected to measure the complementary terminal variables which are necessary to describe the physical behavior of the mechanism. The complementary terminal variables in mechanical systems are the terminal across (translational and rotational velocities) and the terminal through (forces and moments) variables.

The oriented graph representation of the turning-pair connection and the gear-pair connection are shown in Figure 4.

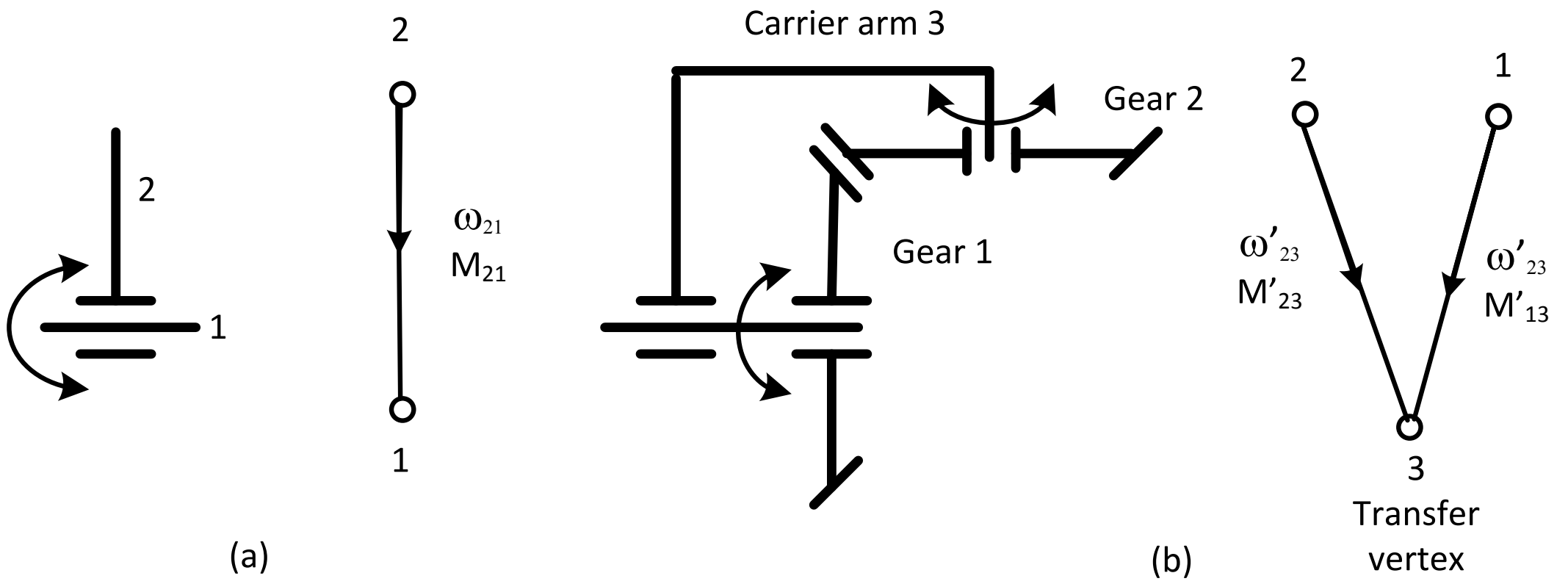

Figure 4: (a) Turning-pair and graph representation, (b) Gear-Pair and graph representation

The relation between the relative velocities and moments of the gears shown in Figure 4(b) is:

$$\begin{bmatrix} \omega'_{13} \\ M'_{23} \end{bmatrix} = \begin{bmatrix} & \mp n_{21} \\ \pm n_{21} & \end{bmatrix} \begin{bmatrix} M'_{13} \\ \omega'_{23} \end{bmatrix} \qquad (26)$$

Where $n_{21} = N_2 / N_1 = 1 / n_{12}$ and $N_1$ and $N_2$ are teeth numbers of gear 1 and gear 2, respectively. The sign of the gear ratio $n$ is determined based on the rotation directions of the gears. If both of the gears rotate in the same direction the sign is (+), otherwise the sign is (-).

In order to obtain the T-T graph representation of the GRM shown in Fig. 3, first, turning pairs are drawn by replacing them with their graph representations. The turning pairs axes labels are inserted into the graph as shown in Fig. 5.

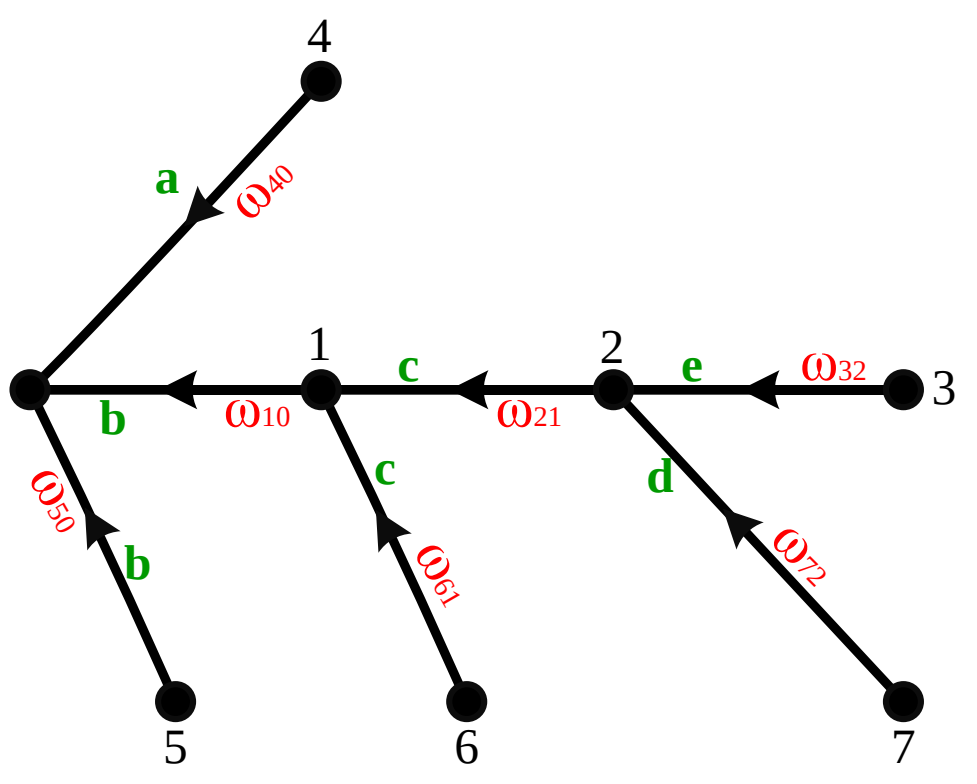

Figure 5: T-T Graph representation of the turning pairs of the GRM.

As it is seen from Fig.5, the turning pairs constitute the tree branches.

Then, the graph representations of the gear pairs are drawn for completing the graph. At this point, it is required to determine the transfer vertices representing the carrier arms for the gear pairs (4,1), (5,2), (6,7) and (7,3). In order to determine the transfer vertex, we will start from one of the node representing the gear in meshes and go through the tree branches to reach the other node representing the other gear. The vertex on this path, which has different levels on opposite sides is the transfer vertex.

- Path 1 $(4 \xrightarrow{a} 0 \xrightarrow{b} 1)$ : vertex 0 (pair axes *a*, *b*),
- Path 2 $(5 \xrightarrow{b} 0 \xrightarrow{b} 1 \xrightarrow{c} 2)$ : vertex 1 (pair axes *b*, *c*),
- Path 3 $(6 \xrightarrow{c} 1 \xrightarrow{c} 2 \xrightarrow{d} 7)$ : vertex 2 (pair axes *c*, *d*),
- Path 4 $(7 \xrightarrow{d} 2 \xrightarrow{e} 3)$ : vertex 2 (pair axes *d*, *e*).

Therefore the sets of gear pair and corresponding carrier arm are (4,1)(0), (5,2)(1), (6,7)(2) and (7,3)(2).

Figure 6 shows the T-T graph representation of the GRM. The thin lines representing the gear meshes constitute the co-tree branches or links.

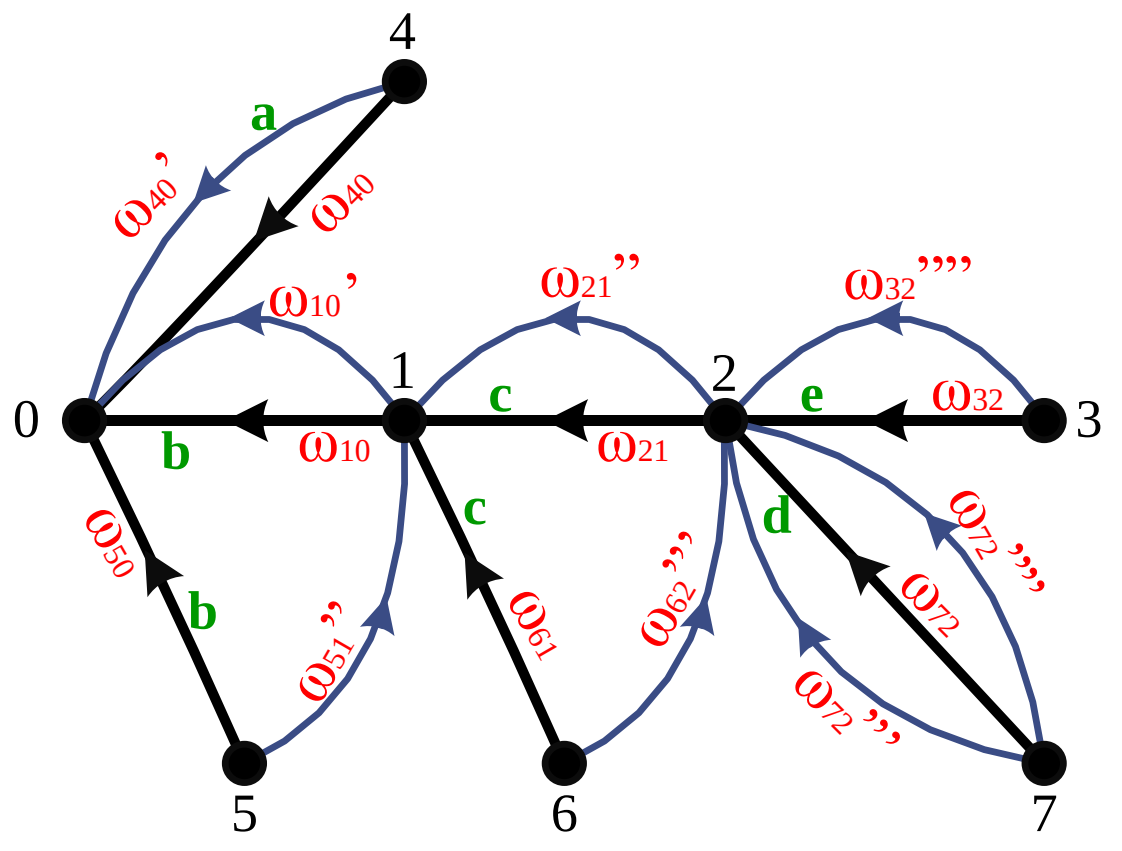

Figure 6. T-T graph representation of the GRM.

For the kinematic analysis of the GRM, we will consider only the angular velocities. Therefore the following terminal equations can be written by using Eq.(26) for the gear pairs:

$$(4,1)(0): \omega'_{40} = -n_{14}\omega'_{10} \tag{27}$$

$$(5,2)(1): \omega''_{51} = -n_{25}\omega''_{21} \tag{28}$$

$$(6,7)(2): \omega'''_{62} = -n_{76}\omega'''_{72} \tag{29}$$

$$(7,3)(2): \omega''''_{72} = n_{37}\omega''''_{32} \tag{30}$$

From the graph shown in Fig. 5, the following fundamental circuits (f-circuit) equations can be written easily.

$$\omega''_{51} = \omega_{50} - \omega_{10} \tag{31}$$

$$\omega'''_{62} = \omega_{61} - \omega_{21} \tag{32}$$

$$\omega'_{40} = \omega_{40} \tag{33}$$

$$\omega'_{10} = \omega_{10} \tag{34}$$

$$\omega''_{21} = \omega_{21} \tag{35}$$

$$\omega'''_{72} = \omega_{72} \tag{36}$$

$$\omega''''_{72} = \omega_{72} \tag{37}$$

$$\omega''''_{32} = \omega_{32} \tag{38}$$

Then, unknown angular velocities can be determined in terms of input velocities by using the terminal and f-circuit equations.

The angular velocities $\omega_{40}, \omega_{50}$, and $\omega_{61}$ are inputs and $\omega_{10}, \omega_{21}, \omega_{32}$, and $\omega_{72}$ are unknown angular velocities for the GRM.

Using Eqs.(33), (33), and (27) $\omega_{10}$ can be obtained as:

$$\omega'_{10} = \omega_{10} = -1/n_{14}\omega'_{40} = -n_{41}\omega_{40} \tag{39}$$

Eqs.(35), (28), and (39) yields:

$$\omega_{21} = \omega''_{21} = -n_{52}\,\omega''_{51} = -n_{52}\left(\omega_{50} - \omega_{10}\right) = -n_{52}n_{41}\omega_{40} - n_{52}\omega_{50} \tag{40}$$

For $\omega_{32}$, Eqs.(38), (30), (37), (36), (29), (32), and (40) are used.

$$\begin{aligned} \omega_{32} &= \omega'''''_{32} = n_{73}\,\omega'''''_{72} = n_{73}\omega_{72} = n_{73}\left(-n_{67}\,\omega'''_{62}\right) = -n_{73}n_{67}\left(\omega_{61} - \omega_{21}\right) \\ \omega_{32} &= n_{73}n_{67}\omega_{21} - n_{73}n_{67}\omega_{61} = -n_{73}n_{67}n_{52}n_{41}\omega_{40} - n_{73}n_{67}n_{52}\omega_{50} - n_{73}n_{67}\omega_{61} \end{aligned} \tag{41}$$

Finally, $\omega_{72}$ can be obtained from Eq.(41):

$$\omega_{72} = n_{37}\omega_{32} = -n_{67}n_{52}n_{41}\omega_{40} - n_{67}n_{52}\omega_{50} - n_{67}\omega_{61} \tag{42}$$

Or in more compact form:

$$\begin{pmatrix} \omega_{10} \\ \omega_{21} \\ \omega_{32} \\ \omega_{72} \end{pmatrix} = \begin{bmatrix} -n_{41} & 0 & 0 \\ -n_{52}n_{41} & -n_{52} & 0 \\ -n_{73}n_{67}n_{52}n_{41} & -n_{73}n_{67}n_{52} & -n_{73}n_{67} \\ -n_{67}n_{52}n_{41} & -n_{67}n_{52} & -n_{67} \end{bmatrix} \cdot \begin{pmatrix} \omega_{40} \\ \omega_{50} \\ \omega_{61} \end{pmatrix} \tag{43}$$

If we use the same notations for gear ratios as shown in Eq.(20) by using the following relations

$$n_{41} = i_{15},\ n_{52} = i_{16},\ n_{67} = i_{17} \text{ and } n_{73} = i_{18} \tag{44}$$

Then the following equation is obtained which is the same as Eq.(25)

$$\begin{pmatrix} \omega_{10} \\ \omega_{21} \\ \omega_{32} \\ \omega_{72} \end{pmatrix} = \begin{bmatrix} -i_{15} & 0 & 0 \\ -i_{16}i_{15} & -i_{16} & 0 \\ -i_{18}i_{17}i_{16}i_{15} & -i_{18}i_{17}i_{16} & -i_{18}i_{17} \\ -i_{17}i_{16}i_{15} & -i_{17}i_{16} & -i_{17} \end{bmatrix} \cdot \begin{pmatrix} \omega_{40} \\ \omega_{50} \\ \omega_{61} \end{pmatrix} \tag{45}$$

**Conclusion**

In this paper, the kinematic equations of the GRM are obtained by using Matroid and T-T Graph Methods in sequence. Both methods use oriented graphs and represent the links with nodes. In both methods, representations of turning pairs constitute the tree branches and gear pair representations constitute the links are co-tree branches. Matroid method uses the oriented lines in order to obtain the incident and path matrices and cycle-basis matrix is derived from these two matrices. Then using screw theory kinematic equations of the mechanism is obtained. On the other hand, T-T graph carries more information than Matroid. Since each line is a part of terminal graph of either turning of gear pair it carries complementary terminal variables. Therefore using terminal equations of the gear pairs, fundamental circuit and fundamental cut-set equations kinematic and static equations of the mechanism can be obtained. This example shows that kinematic equations of the GRM can be obtained more easily than Matroid Method by using T-T Graph method. The only advantage of the Matroid method is the derivation of the relative velocities of the gears with respect to carrier arm. Since it uses the diameter of the gears and the diameters are written relative to the reference frame, the sign of the gear ratio is obtained directly without considering the turning direction of the gears.